\documentclass[pdflatex,journal]{journal}
\ifCLASSINFOpdf
\else
\fi

\usepackage[pdftex]{graphicx}
\usepackage[square,numbers,sort&compress]{natbib}
\usepackage{amsmath}
 \usepackage{mathptmx} 

\usepackage[none]{hyphenat}

\usepackage[justification=centering]{caption}

\usepackage[labelsep=space]{caption}

\usepackage[font=footnotesize]{caption} 

\begin{document}
\bstctlcite{IEEEtran} 
%
\title{An Automatic Control System with Human-in-the-Loop for Training Skydiving Maneuvers: Proof-of-Concept Experiment}
%
%
%
\author{Anna~Clarke,~Per-Olof~Gutman
\thanks{A. Clarke is with the Technion Autonomous Systems Program, Technion - Israel Institute of Technology, Haifa 32000, Israel,
              ORCID iD: 0000-0003-3491-3847 
              }
\thanks{P.O. Gutman is with the Faculty of Civil and Environmental Engineering, Technion - Israel Institute of Technology, Haifa 32000, Israel,
         ORCID iD: 0000-0002-6666-0971}
}

\maketitle
\thispagestyle{empty}

\begin{abstract}
A real-time motion training system for skydiving 
is proposed. Aerial maneuvers are performed by changing the body posture and thus deflecting the surrounding airflow. The 
natural learning process is extremely slow due to unfamiliar free-fall dynamics, stress induced blocking of kinesthetic feedback, and complexity of the required movements. 
The key idea is to augment the learner with an automatic control system that would be able to perform the trained activity if it had direct access to the learner’s body as an actuator. The aiding system will supply the following visual cues to the learner: 1. Feedback of the current body posture; 2. The body posture that would bring the body to perform the desired maneuver; 3. Prediction of the future inertial position and orientation if the body retains its present posture. The system will enable novices to maintain stability in free-fall and perceive the unfamiliar environmental dynamics, thus accelerating the initial stages of skill acquisition.
This paper presents  results of a Proof-of-Concept experiment, whereby humans controlled a virtual skydiver free-falling in a computer simulation, by the means of their bodies. This task was impossible without the aiding system, enabling all participants to complete the task at the first attempt. 
\end{abstract}


%
\IEEEpeerreviewmaketitle

\section{Introduction}
\label{intro_literature}
%
%
%
%
\IEEEPARstart{F}{ree} fall maneuvering requires maintaining a desired angular and linear velocity, and is achieved by continuously changing the body posture. Novices experience hard and protracted training due to 1. Exteroceptive sensory overload, with the external world  changing at 220 km/h; 2. Blocking of kinesthetic feedback, due to muscle tension caused by stress; 3. The need to break the habitual ways of moving, noting that free fall maneuvering requires counter-intuitive movements, 
different from our daily movement repertoire. 
Moreover, it is impossible to demonstrate these movements since they are highly dependent on individual body parameters, involving the whole body with multiple Degrees-of-Freedom (DOFs). The trainees are taught only a few basic postures.
All current training techniques are off-line: debriefing the jump videos, practicing the basic postures on the ground, and visualizing future maneuvers. 

A new training method is needed due to the rapid growth of skydiving operations along with the number of fatalities \cite{web_USPA}, caused by novice skydivers lacking  control. 

The training tool described here constitutes a hierarchical control system composed of a human performer, an autonomous system capable of performing the activity in a virtual way, and an interface between them. The strengths of control engineering are complementary to human motor-learning abilities.  In many areas humans outperform computer algorithms, namely knowledge-based tasks, domains  requiring expertise, tasks with high levels of uncertainty \cite{cummings2014man}. For sensory-motor actions, however, humans need a training period until they become automatic, particularly when acting in unfamiliar and stressful environments. Motor skills do not improve until the  brain learns the environmental dynamics. Control engineering is a powerful tool for modelling complex dynamics, and designing controllers for maneuvering in challenging environments. In this work those strengths are turned into motor learning aids. 

The proposed approach is conceptually different from a state-of-the-art semi-autonomous and human-machine systems. In such systems humans take a supervisory role over an automated process \cite{wickens2015engineering}, \cite{cummings2014man}. In the proposed training tool the roles are reversed: the control system will guide human movements. The proposed control system will resolve on-line state and parameters estimation, plan the desired trajectory in   3D space, break    down the trajectory  into a series of maneuvers, track  the linear and angular motion involved in each maneuver, and interpret  the controller commands in terms of   body posture. The body actuation will be the manual part: a human trainee will focus on making his current posture close to the one recommended by the control system. 

Motor skills for physical education and sports are  investigated in \cite{button2012empirical}, \cite{schollhorn2012nonlinear}, with theoretical teaching guidelines in \cite{motor_tani2014adaptive}. 
Virtual and Augmented Reality technologies \cite{vr_anderson2013youmove}-\nocite{vr_bergamasco2012skill}\nocite{vr_levin2010feedback}\cite{vr_zimmerli2009virtual} noticeably improved  training by providing feedback, which is an essential motor-learning aid 
\cite{vr_bergamasco2012skill}, \cite{sport_brodie2012dance}-\nocite{sport_diamant1991mind}\nocite{sport_holmes2001pettlep}\cite{motor_wolpert2011principles}. Movement feedback has become available via miniature wireless, wearable inertial sensors,  and movement reconstruction from images.
Relevant, easy-to-perceive, correctly timed feedback improves performance.

State-of-the-art sports training systems include: Tai-Chi gestures \cite{chua2003training}, \cite{portillo2010capturing}, martial art training \cite{kwon2005combining}, golf swing \cite{ghasemzadeh2009wearable}, dance and other recorded moves  
\cite{yang2002implementation}, \cite{baek2003motion}, 
\cite{vr_anderson2013youmove}, \cite{portillo2010capturing}, 
all based on imitation: specific, precisely known moves are repeated after a coach/avatar through a virtual reality interface and the disparity in execution  is the feedback. Feedback is supplied as a score, via superimposition of   correct and   trainee limbs trajectory, or via a multimodal (visual, vibrotactile, sound) interface  \cite{vr_bergamasco2012skill}. 
The question remains as to how to train in sport for which imitation is impossible. In skydiving posture adjustments happen very fast, can be barely noticeable but vital, simultaneous in all body limbs, and individual: templates of  'correct' moves don't exist. 
An elegant solution for   real-time feedback was proposed in \cite{spelmezan2012investigation},  \cite{moeyersons2016biofeedback}  for novice snowboard training, and in \cite{kirby2009development} for expert alpine ski training. Since these activities are cyclic  
a characteristic dynamic variable  for feedback was found: for snowboarding  the weight transfer given as  tactile stimulation 
\cite{spelmezan2012investigation}; and for skiing  the lateral displacement during each turn given as audio feedback \cite{kirby2009development}. 

In free-fall motion is not constrained by the ground, all  limbs are free to move as desired. Their slightest movement can cause large aerodynamic forces/moments that can produce uncontrollable body rotation and horizontal displacement with continuously increasing speed, something typical for students. The  challenge  is thus to acquire  understanding and  feeling of the environment dynamics, 
called the \textit{Forward Model} in motor learning literature. This model allows the brain to predict the motion caused by different body postures, and issue  desired posture adjustments to the somatic nervous system.  However, acquisition of the Forward Model occurs only by actively moving inside the new environment and processing the sensory feedback \cite{sport_diamant1991mind}.   
Hence, the difficulty in skydiving training is moving in free-fall before the body has developed the necessary \textit{movement patterns}: combinations of body DOFs activated synchronously and proportionally, as a single unit. 

A sports technique can be viewed as a movement patterns repertoire. 
From the perspective of dynamical systems theory (the dominant motor learning theory), motor learning is the process when these movement patterns emerge \citep{sport_schmidt2008motor}. First, the patterns are simple (coarse), providing just the basic functionality. Later, the movement patterns become more complex, providing adaptation to perturbations and uncertainties, and improved performance. However, in skydiving it is extremely difficult to control the motion using simple patterns,  taught to novices, due to the unstable and highly non-linear aerodynamics of free-fall,  shown in \cite{article}. Students can spend most of their free-fall training session  attempting to regain stability, i.e. stop an undesirable and often dangerous motion.     
  
This vicious circle can be broken introducing our system \cite{clarke2018computerized}, called Kinesthetic Training Module (KTM), into training. It supplies to the trainee the outcome of the missing Forward  Model: the prediction of motion in  inertial space. It also shows the  control input required at each time instant  in terms of the movement pattern being trained. Each exercise suggested by KTM is based on the student's current movement repertoire  which is continuously monitored and analysed 
\cite{patterns_analysis}. Students are given achievable tasks: desired maneuvers are outputs of simulations driven by trained patterns.
\begin{figure}[!t]
     \includegraphics[width=1\columnwidth]{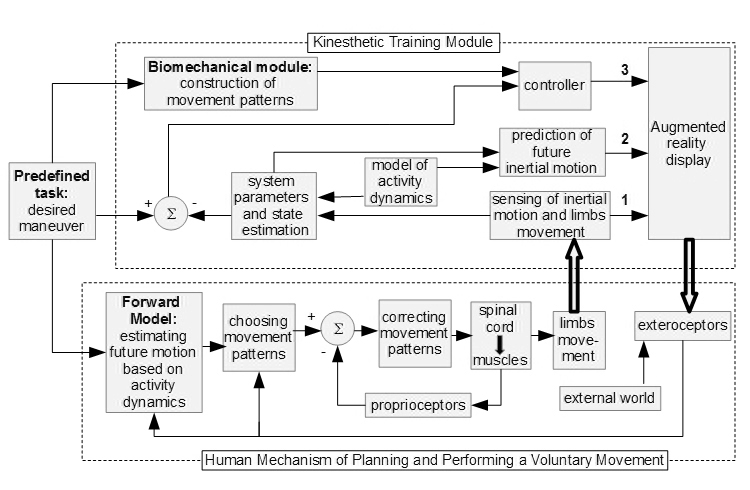}
    \caption{Block Diagram of a natural neurophysiological mechanism augmented with KTM. The cues are denoted as (1) - Feedback Posture, (2) - Forward Model, (3) - Desired Posture}
    \label{fig:diagKTM}
\end{figure}

\section{Methods}
The experiments were approved by the Technion Ethics Committee. All participants expressed a written informed consent.
\subsection{Concept Outline}
\label{concept}
KTM computes \textit{Feedback Posture}, \textit{Forward Model}, and \textit{Desired Posture}, and displays them in real-time, as shown in Fig. \ref{fig:diagKTM} together with a block diagram representing human movement, as explained in Sec.\ref{intro_literature}.
KTM includes three hardware components: a sensor suite for tracking body movements, augmented reality goggles for displaying the cues, and a wearable computer for real-time computations.
  
The  Forward Model provides future inertial position and orientation if the  current posture is not altered. This cue functions like feed-forward in control theory. Its computation requires modeling   body and environment dynamics, estimating system parameters and state variables, and solving the motion equations forward in time. The prediction time has to match the human sensory-motor bandwidth, to be found and tuned experimentally, individually for each trainee.  
The Skydiver Simulator, developed for this purpose, includes the modules briefly described below, while the exact equations can be found in \cite{article}.

\paragraph{Biomechanical Model} represents the body by 16 rigid segments (pelvis, abdomen, thorax, head, upper arms, forearms, hands, upper legs, lower legs, and feet) of simple geometrical shapes and calculates the local centre of gravity and principal moments of inertia for each segment. A set of rotation quaternions linking each two segments enables computation of the overall centre of gravity, inertia tensor, and their time derivatives. The model has altogether 45 DOFs: 3 rotation angles associated with each one of the 15 joints. These DOFs define an instantaneous body posture, which is the simulator's input. The biomechanical model has to be provided with anthropometric parameters specifying body size, shape, and weight of the skydiver under investigation. 
\paragraph{Dynamic Equations of Motion} derived by the Newton-Euler method, provide six equations: 3D forces and moments. 
\paragraph {The Kinematic Model} computes the body inertial orientation, and angles of attack, sideslip, and roll of each segment relative to the airflow. These angles are used in the aerodynamic model to compute drag forces and aerodynamic moments acting on each segment. The total aerodynamic force and moment together with the gravity forces are substituted into the equations of motion. 
\paragraph{Aerodynamic model} is formulated as a sum of forces and moments acting on each individual segment, modelled similar to aircraft aerodynamics - proportional to velocity squared and to the area exposed to the airflow. The model includes six aerodynamic coefficients (maximum lift, drag, and moment coefficients; roll, pitch, and yaw damping moment coefficients) that were experimentally estimated, and has to be provided a set of configuration parameters specific to the skydiver under investigation (type of parachute, helmet, jumpsuit, and weight belt). 

The skydiving simulator output was experimentally verified: Various skydiving maneuvers were performed by different skydivers in a wind tunnel and in free-fall, and the recorded posture sequences were fed into the simulator. The six aerodynamic coefficients were tuned so that all the manoeuvres were closely reconstructed. The errors RMS in angular and linear velocities were 0.15 [rad/sec] and 0.45 [m/sec], respectively. 

\begin{figure}[!t]
     \includegraphics[width=1\columnwidth]{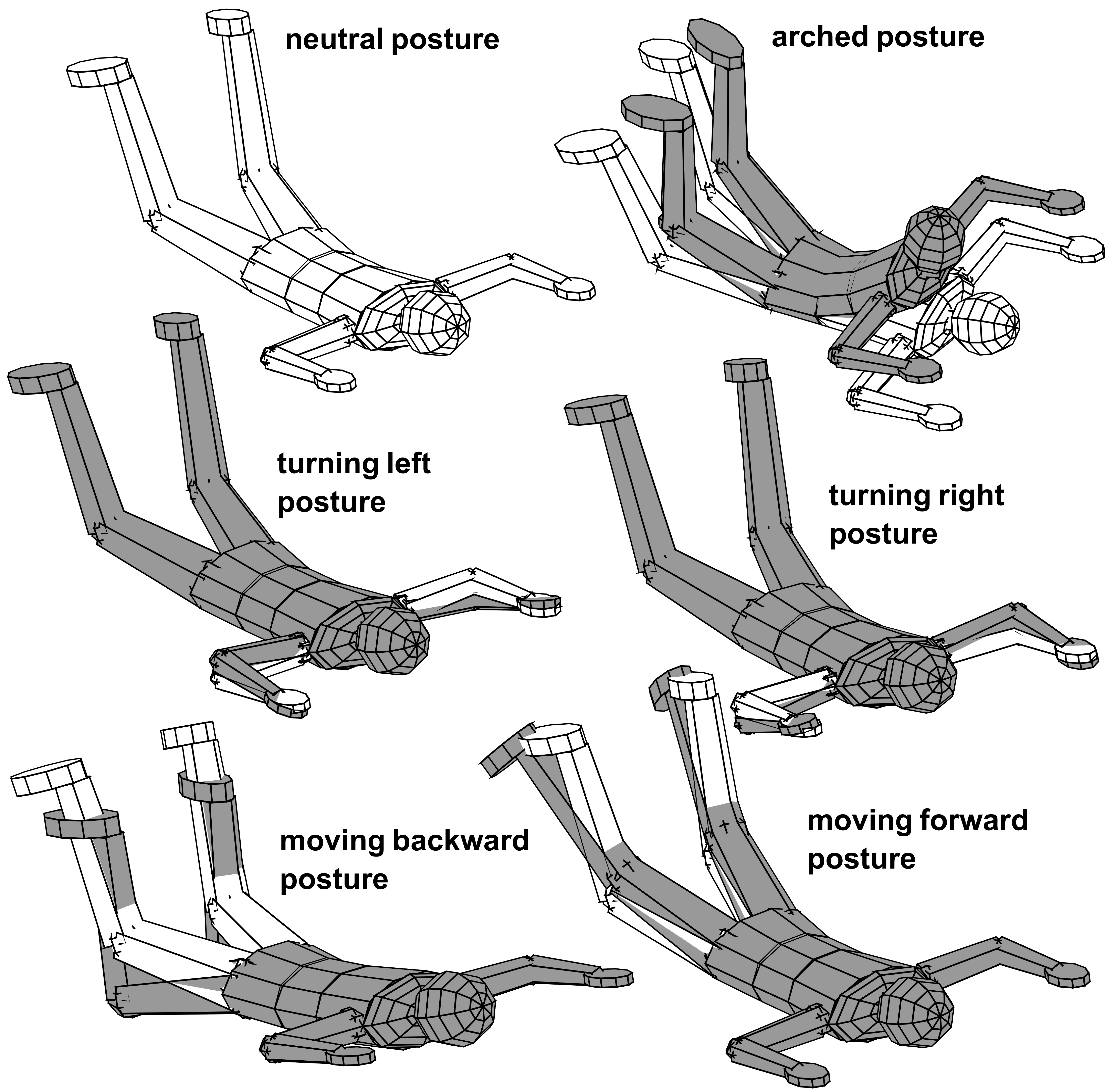}
    \caption{The neutral skydiving posture and other postures (dark) superimposed over the neutral one. }
    \label{fig:pos}
\end{figure}
 
The Desired Posture, constructed from the specific movement patterns currently being trained, superimposed on the Feedback Posture provides a cue of required instantaneous posture adjustments. 
Movement patterns for training can be generated from empirical databases of skydiving experts, and finalized with the Skydiver Simulator, such that each movement pattern has a good performance potential from control engineering perspective.  For example, in \cite{patterns_analysis} the trainee's movement pattern is adjusted in the simulator such that the overall open loop system (i.e. body in free-fall actuated by the refined movement pattern) acquires dynamic stability.  
For each movement pattern a controller is designed that tracks a matching component of the linear or angular velocity  associated with the desired maneuver. The Desired Posture is thus a superposition of active controller outputs, see \cite{MEDarticle} where three controllers tracking longitudinal, lateral, and vertical motion define the final body posture command. In the general case the Desired Posture at every instant of time $P_{desired}(t)$ is computed as:
\begin{equation}
P_{desired}(t)=P_{neutral} + \sum_{i=1}^{N}u_i(t) \cdot MP_i
\end{equation}
where $P_{neutral}$ is the basic skydiving posture used to fall straight down in a belly-to-earth position (see Fig. \ref{fig:pos}), $N$ is the total number of movement patterns involved in constructing the Desired Posture, $u_i(t)$ is the command computed by controller $i$ at time $t$, and $MP_i$ is the eigenvector (with norm 1) defining movement pattern $i$: the 45 components of this vector define the relative engagement of each body DOF. 

Notice, that for efficient training $P_{neutral}$ and $MP_i, \quad 1\leq i \leq N$ have to be designed individually for each trainee due to different anthropometrics. The input to each controller is the discrepancy between the relevant desired and measured velocity component (e.g. yaw rate error), and the output is the pattern angle in radians $u_i(t)$. Each controller is designed by a method most suited for dealing with the dynamics of the tracked motion. For example, in \cite{MEDarticle} a controller tracking the vertical velocity component is designed according to Model Predictive Control (MPC): a method that deals with state and actuation constraints. The reason is that the vertical velocity in free-fall is adjusted by arching ones back (as shown in Fig. \ref{fig:pos}), which has a very limited range for most people.    
In case a pre-defined trajectory is desired, linear and angular velocity profiles are computed by a path-following controller \cite{article}. 
The desired velocities can enhance the Forward Model cue: the trainee can compare his future motion state with the state intended by the control system. If he implements precisely the desired posture, his future and planned motion will coincide.

Training practice starts with simple movement patterns, involving few DOFs. Gradually, the trained patterns become more complex, and more patterns can be practiced simultaneously. The trainee can be suggested other patterns for already mastered maneuvers, using different DOFs. This possibility, granted by the natural human kinematic redundancy, can be highly beneficial for accelerating the skill acquisition \cite{motor_tani2014adaptive}. Theoretically, the KTM cues contain sufficient information to trigger the emergence of more complex  patterns based on the combination of simpler, previously trained ones \cite{Qlearning}. The coach can continuously monitor the progress, detect the emergence of particularly efficient patterns, and trigger their training  in the next sessions, as shown in \cite{patterns_analysis} for intermediate-level rotation training. Designing and adapting exercises to different types of learners is in the main stream of our future work.   
    
We hypothesize that mastering maneuvers rapidly and maintaining dynamic stability will be best achieved if the subject focuses on causing the current body posture and the predicted inertial motion to coincide with the displayed cues, respectively. 
Due to the perception-action coupling we expect each learning step  
to include two sub-steps: After some training the pattern becomes part of the muscle memory, the posture cue becomes redundant, and the Forward Model cue will be in focus. When the next pattern is practiced, the focus will return to the Desired Posture cue, etc. 
When also the Forward Model cue becomes unnecessary, the environmental dynamics has been learnt, and the trainee can  perform the maneuver without the KTM. 
     
\subsection{Proof-Of-Concept Experiment}

The KTM concept suggests that the trainee, guided by the visual cues, is part of the control loop. The trainee is the actuator: the real-time controllers compute a body posture command for the trainee to execute. The proof-of-concept experiment is meant to verify that one may design a stable hierarchical control system where the human implements the computed continuously changing body posture. 
Verifying the effectiveness of the KTM cues and display, and developing KTM-prototype guidelines are additional objectives.

In \cite{article} a navigation and control algorithm for a virtual skydiver was developed. It enabled a skydiver in the simulation to fly from his current location to reach another skydiver. The navigation algorithm planned a path connecting the virtual skydiver to his target, and computed the desired yaw rate and speed profiles. The controller interpreted these profiles as two commands: 'turning' and 'forward-backward' movement pattern angles, respectively. These two patterns (see Fig. \ref{fig:pos}) defined the posture at each time instant. The 'turning' pattern, used for tracking the yaw rate, included four DOFs associated with the shoulders. The 'forward-backward' pattern, used for tracking the speed, included four DOFs associated with knees and hips. 

The same task is used for our Proof-Of-Concept experiment. In contrast to the simulated autonomous skydiver, a computed body posture is executed by the trainees who view the skydive simulation on a screen in real-time. Thus, the Desired Posture cue displayed to the trainees is computed as: 
\begin{equation}
P_{desired}(t)=P_{neutral} + u_{arms}(t) \cdot MP_{arms}+ u_{legs}(t) \cdot MP_{legs}
\end{equation}
where $MP_{arms}$ is the 'turning' movement pattern eigenvector with 4 non-zero entries $MP_{arms}(i_1,i_2,i_3,i_4)=0.5$, with $i_1-i_4$ associated with right shoulder flexion and lateral rotation, and left shoulder extension and medial rotation; $MP_{legs}$ is the 'forward-backward' movement pattern eigenvector with 4 non-zero entries $MP_{legs}(j_1,j_2)=0.582; \quad MP_{legs}(k_1,k_2)=0.402$, with $j_1,j_2$ associated with the knees flexion and $k_1,k_2$ with hips extension; and $u_{arms}(t)$, $u_{legs}(t)$ are controller's commands computed as:
\begin{equation}
\begin{split}
u_{arms}(t) = &G11 \cdot (F11 \cdot \Omega_{com}(t)-\Omega_{meas}(t))\\
u_{legs}(t) = &G22 \cdot (F22 \cdot V_{com}(t)-V_{meas}(t))+...\\
&G21 \cdot (F11 \cdot \Omega_{com}(t)-\Omega_{meas}(t))
\end{split}
\label{eq:contr}
\end{equation}
where $\Omega_{com}(t)$, $V_{com}(t)$ are the yaw rate and speed commands at time $t$;  $\Omega_{meas}(t)$, $V_{meas}(t)$ are the skydiver's yaw rate and speed at time $t$ computed by the Skydiving Simulator that receives the measured trainee's posture at 240 [Hz] and propagates in time the equations of motion; and $F11, F22, G11, G12, G22$ define the control law and are given as: 
\begin{equation}  \label{eq:loop1}
\begin{split}
&G_{11}=\frac{0.25(1+\frac{s}{3.5})}{s} \frac{1+\frac{s}{0.7}}{1+\frac{s}{10}} \frac{1}{1+\frac{s}{100}}\\
&F_{11}=\frac{1}{(1+\frac{s}{7})(1+\frac{s}{8})}  \frac{1+\frac{s}{0.6}}{1+\frac{s}{1}} \\
&G_{22}=\frac{0.1(1+\frac{s}{1.5})}{s} \frac{1+\frac{s}{0.2}}{1+\frac{s}{0.6}} \frac{1+\frac{s}{1}}{1+\frac{s}{10}}\\
&G_{21}=\frac{-0.035(1+\frac{s}{3})}{s} \frac{1+\frac{s}{1}}{s} \frac{1+\frac{s}{0.5}}{1+\frac{s}{5}} \\
&F_{22}=\frac{1}{(1+\frac{s}{1})(1+\frac{s}{2})}
\end{split}
\end{equation}
where $s$ is the Laplace variable The controller was designed using the Quantitative Feedback Theory (QFT)  providing  robustness  against  plant  non-linearities  and  inaccurate  execution of the movement patterns. The design procedure is described in \cite{article}.

Notice, that the yaw rate and speed commands $\Omega_{com}(t)$, $V_{com}(t)$ in Eq. \ref{eq:contr} are not the same as the pre-planned desired yaw rate and speed profiles, which are computed offline along with the desired path and are a part of the task definition. The reason is that the real time yaw rate and speed commands have to account for heading and position errors produced by the trainee during task execution. Therefore, these commands are computed as:
\begin{equation}
\begin{split}
V_{com}(t) &= V_{desired}(t,t_{LA}) \\
\Psi_{error}(t) &= atan \frac{Xpath(t,t_{LA})-X(t)}{Ypath(t,t_{LA})-Y(t)}-atan\frac{V_x(t)}{V_y(t)}\\
\Omega_{com}(t) &= 2 \cdot \Psi_{error}(t)/t_{LA}
\end{split}
\label{eq:eq_control}
\end{equation}
where $X(t), Y(t)$ is the horizontal position of the skydiver at time $t$; $V_x(t), V_y(t)$ are his horizontal velocity components; $t_{LA}$ is the look-ahead time conveying the reaction time expected from the trainee, it can be individually adjusted and was between 2-2.5 seconds in our experiments; $Xpath(t,t_{LA}), Ypath(t,t_{LA})$ is the path point located at a distance $t_{LA}\cdot \sqrt{V_x(t)^2+V_y(t)^2}$ from the path point closest to $X(t), Y(t)$; and $V_{desired}(t,t_{LA})$ is the speed from the desired speed profile corresponding to the path point  $Xpath(t,t_{LA}), Ypath(t,t_{LA})$. 

Trainees have to learn, firstly, to reproduce the 'turning' and 'forward-backward' patterns, and secondly, to use these patterns for flying the virtual skydiver. The Proof-of-Concept is a success if all the trainees 
complete the task after a small number of attempts. 
Without the KTM cues it is impossible to control the virtual skydiver. The reason is not only  the unfamiliar environmental dynamics that drives the simulator, but also the total absence of kinesthetic feedback, since the action happens in the virtual world and the teleoperator's proprioceptors cannot feel the air flow, and his vestibular system cannot feel the acceleration and angular rates. Similarly for real skydiving novices: their kinesthetic feedback is blocked by the visual sensory overload and excessive muscle tension due to stress.      
\begin{figure}[!t]
     \includegraphics[width=1\columnwidth]{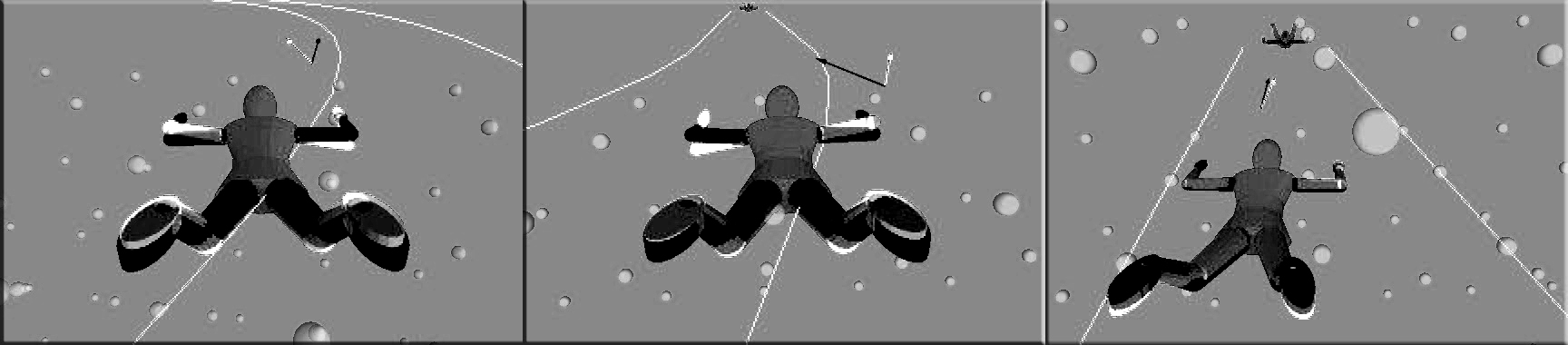}
    \caption{Snapshots of the simulation screen when one of the volunteers performed the task. Light posture and dark arrow are the commands; dark posture and light arrow are the actual posture and future heading of the participant}
    \label{fig:proofexp}
\end{figure}
\subsection{Experimental Set-Up and Procedure}
The Proof-Of-Concept included the following components:
\begin{enumerate}
\item Xsens motion tracking suit \cite{sensor_roetenberg2009xsens} with miniature inertial sensors that are fixed at strategic locations on the body. Each unit includes a 3D accelerometer, 3D rate gyroscope, 3D magnetometer, and a barometer. It has a battery and a small computer located on the back and not restricting the skydiving-specific movements. The Xsens output is wirelessly transmitted to a PC at 240 Hz. Each measurement set includes the orientation of 23 body segments (pelvis, four spine segments, neck, head, shoulders, upper arms, forearms, hands, upper legs, lower legs, feet, toes) relative to the inertial frame, expressed by quaternions. The measurements accuracy is less than 5 degrees RMS of the dominant joint angles. 
\item Skydiver Simulator (PC/Matlab) that receives the measurements via User Datagram Protocol,  computes the skydiver's movement and the KTM cues, and displays the results using the Matlab Virtual Reality Modeling Language. 
\item Volunteers: four women and eight men aged 25-45 years. 
\end{enumerate}
The Experimental Procedure for one trainee at a time consisted of the following stages:
\paragraph{Preparation} The trainee's body parameters are measured and saved in Xsens software; he dresses the Xsens suit, and performs calibration. The calibration is needed for the Xsens internal biomechanical model to converge and includes standing still and walking back and forth for a couple of minutes. Next, the trainee sits on a chair in front of the virtual world display. The trainee is to move his lower legs and arms which is sufficient for this experiment. The remaining  
body is assumed to be in  neutral skydiving posture. The virtual world includes the host and target skydivers and the sky, represented by a grid of stationary white dots, which 
enable the trainee to perceive motion of the host skydiver. The virtual world is viewed from behind and slightly above of the host skydiver.
\paragraph{Introduction} The trainee  is introduced to the simulator and tries to fly the virtual skydiver using his body. After this acquaintance with the simulator, he gets the task to fly towards the target skydiver, without the KTM cues. The results are recorded.   
\paragraph{Training} Planned according to the Explanation, Demonstration, Imitation, and Practice   Method \cite{berry2018becoming} originally developed for military training and also efficiently used for skydiver emergency procedures training. 
\begin{enumerate}
\item \textbf{Explanation:} The two patterns to turn and to move forward/backward are explained to the trainee. The body feeling of the patterns is described, and what one needs to pay attention to. 
In addition explanatory aids were occasionally used: holding a stick emphasizes that the hands should stay in their original position, while the elbows move. Some trainees chose to keep  the stick during the Imitation stage. 
\item \textbf{Demonstration:} The patterns are demonstrated to the trainee, who is immediately asked to move the arms several times between two extreme positions of the 'turning' pattern. This is required for mapping, whereby the simulator can map the 'turning' pattern   in a sitting position to a 'turning' pattern   in an arched free-fall position. 
\item \textbf{Imitation:} After mapping the trainee views himself in a belly-to-earth position in a virtual world and a half-transparent desired posture. Initially the desired posture corresponds to  neutral fall, then continuous right turn, continuous left turn, and, finally, the desired posture  becomes dynamic. The 'turning' pattern is activated in a slow sine wave:\newline $P_{desired}(t)=P_{neutral} + 10\cdot \frac{\pi}{180} \cdot sin(2\cdot \pi \cdot 0.25\cdot t) \cdot MP_{arms}$ \newline
Notice that the maximum desired magnitude of the 'turning' pattern is small: 10 [deg], as can also be seen in Fig. \ref{fig:pos}.
At each stage the trainee is required to make his posture coincide with the desired one such that the error between them stays within a pre-defined threshold for a few seconds. When all the imitating exercises are completed, the desired posture  returns to neutral and the practice begins.  
\item \textbf{Practice:} The purpose   is to use the two   patterns  to fly to the target skydiver. KTM cues are  continuously displayed, see snapshots in Fig. \ref{fig:proofexp}. Upon task completion the trainee is asked to   state which  cue was most helpful, and point out strengths and weaknesses of the user interface. 
\end{enumerate}

\paragraph{KTM Cues} The Feedback Posture is superimposed over the Desired Posture, Fig. \ref{fig:proofexp}, that is displayed half-transparent and in different color. The Desired Posture 
 has a limited range, with each joint  limited to a maximal allowed rotation, and rate of change. Thus the Desired Posture changes slowly enough to be perceived by a human observer, and stays within the ergonomic range of motion. The two arrows   in front of the skydiver in Fig. \ref{fig:proofexp} constitute the Forward Model cue: they show the predicted and the desired position and heading in $t_{predict}$ seconds, where $t_{predict}$  is a tuning parameter, here about 2 seconds. The position of the arrows,  and pointing direction are computed by advancing the kinematic equations  $t_{predict}$~seconds, undert the assumption that the following is known  for the   interval  $[t, t+t_{predict}]$:
 \begin{enumerate}
     \item For the 'desired' arrow - the 'desired' angular velocity from the navigation module $\Omega_{com}(t)$, and the current linear velocity $V_{meas}(t)$, and
     \item For the 'predicted' arrow - the current linear and angular velocities $V_{meas}(t)$ and $\Omega_{meas}(t)$, respectively.
 \end{enumerate}
 The two arrows on screen start from the same point, in order to  enhance their difference in orientation, and thus allow trainees to follow the path at their own pace, not necessarily as prescribed by the pre-planned speed profile. 
 
 If the trainee makes the two arrows coincide, his motion in the coming $t_{predict}$ seconds will be as desired.
\paragraph{The Desired Trajectory or Task}
The Feedback and Desired Postures help to solve the Low Level Control problem:  actuation, i.e. how to cause motion. The arrows representing the Forward Model help to solve the Medium Level Control problem: tracking the linear and angular velocity profiles, i.e. how much to move. One more cue is needed for the High Level Control problem: navigation, i.e. where to move. 
In our case this cue is a path connecting the initial location of the user with the targeted position, represented by two parallel lines, outlining a corridor of desired motion. The easiest and fastest way to the target is following this corridor at every time instant. 

\section{Results}
\begin{figure}[!t]
     \includegraphics[width=1\columnwidth]{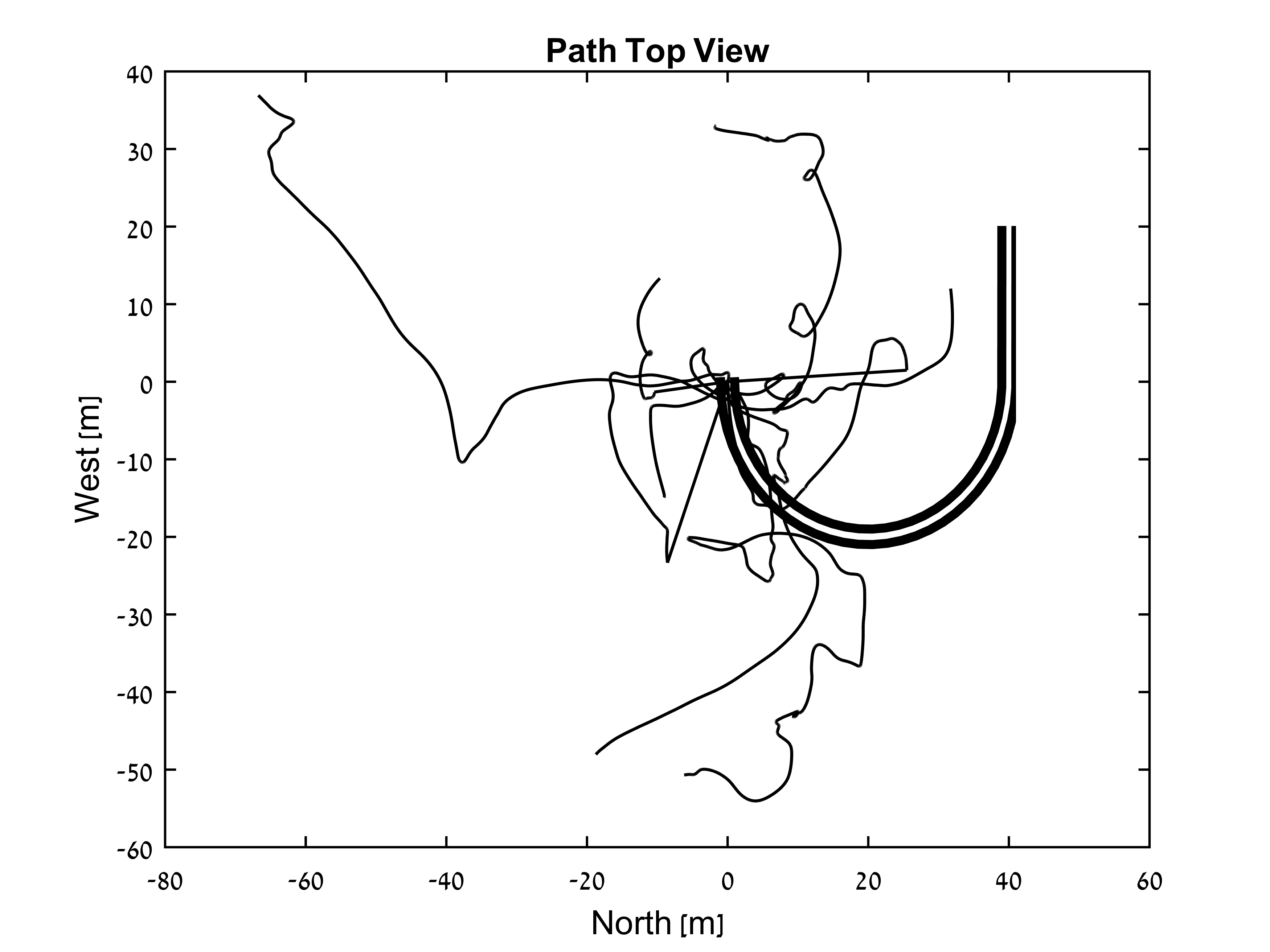}
    \caption{Top view of the desired corridor (thick lines) and actual trajectories of participants 1-6 (thin lines) performing the task without the aid of KTM cues}
    \label{fig:group1_topview_free}
\end{figure}

\begin{figure}[!t]
     \includegraphics[width=1\columnwidth]{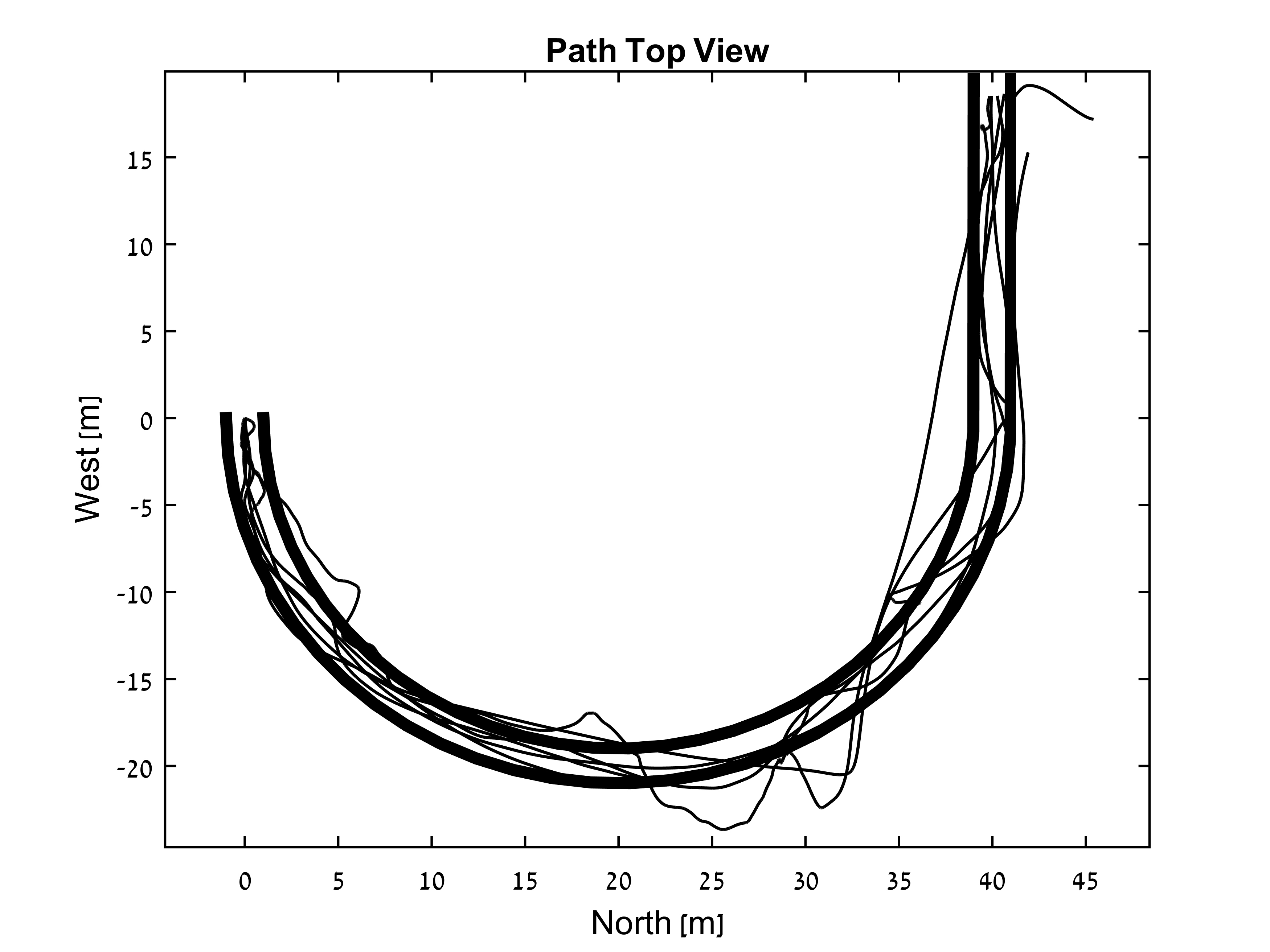}
    \caption{Top view of the desired corridor (thick lines) and actual trajectories of participants 1-6 (thin lines) performing the task with the aid of KTM cues}
    \label{fig:group1_topview}
\end{figure}

\begin{figure}[!t]
     \includegraphics[width=1\columnwidth]{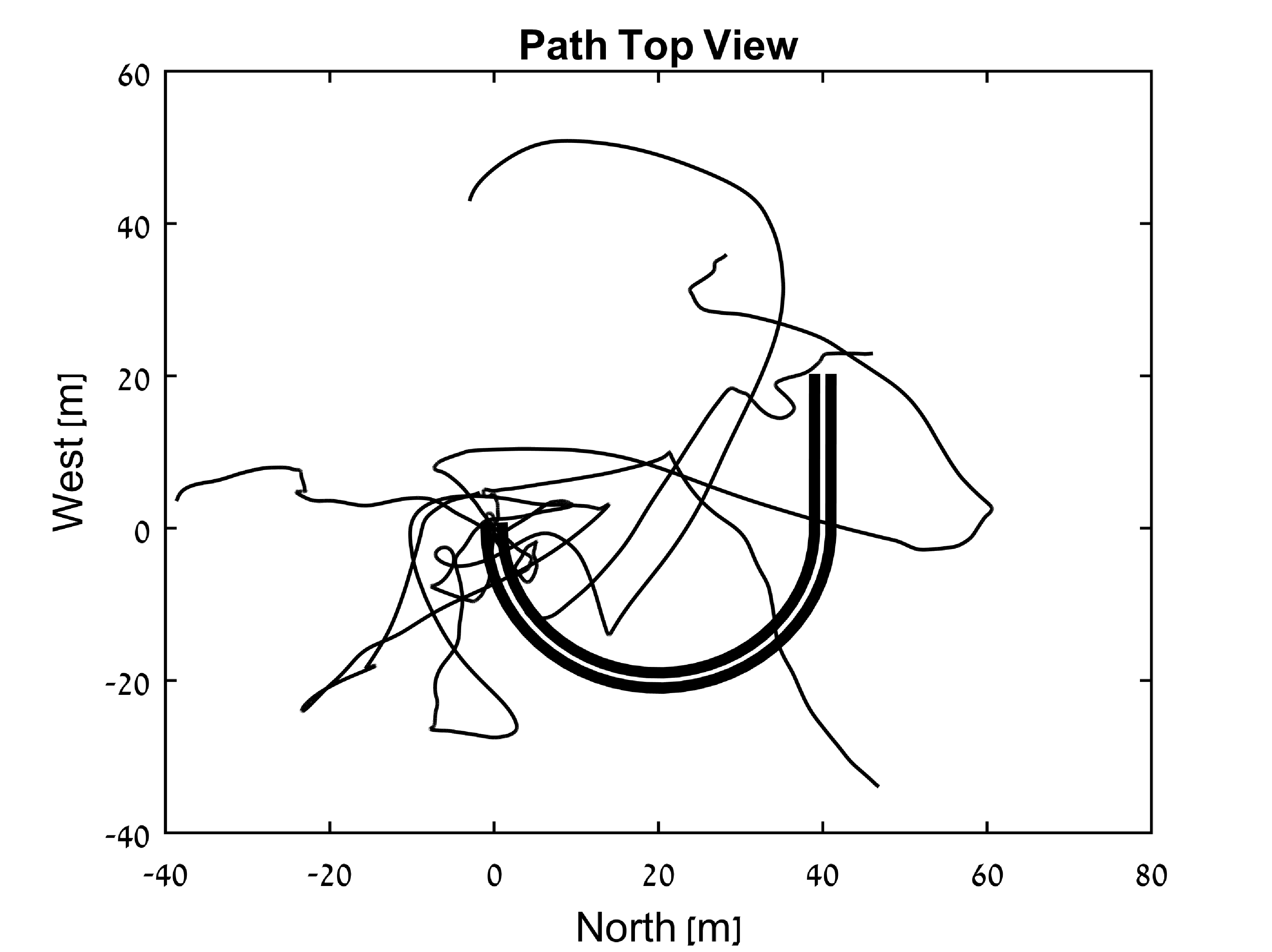}
    \caption{Top view of the desired corridor (thick lines) and actual trajectories of participants 7-12 (thin lines) performing the task without the aid of KTM cues}
    \label{fig:group2_topview_free}
\end{figure}

\begin{figure}[!t]
    \centering
     \includegraphics[width=1\columnwidth]{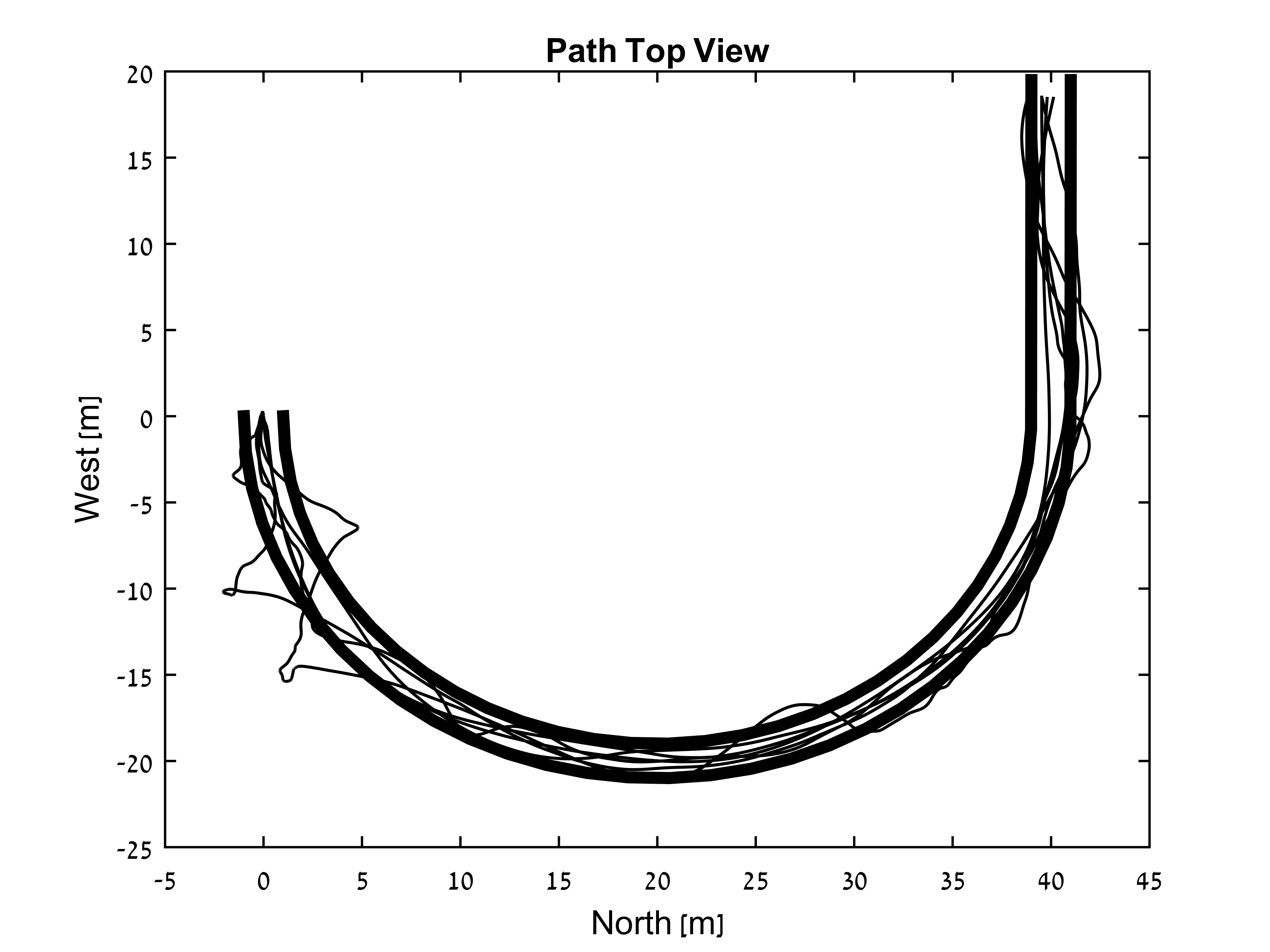}
    \caption{Top view of the desired corridor (thick lines) and actual trajectories of participants 7-12 (thin lines) performing the task with the aid of KTM cues}
    \label{fig:group2_topview}
\end{figure}

\begin{figure}[!t]
     \includegraphics[width=1\columnwidth]{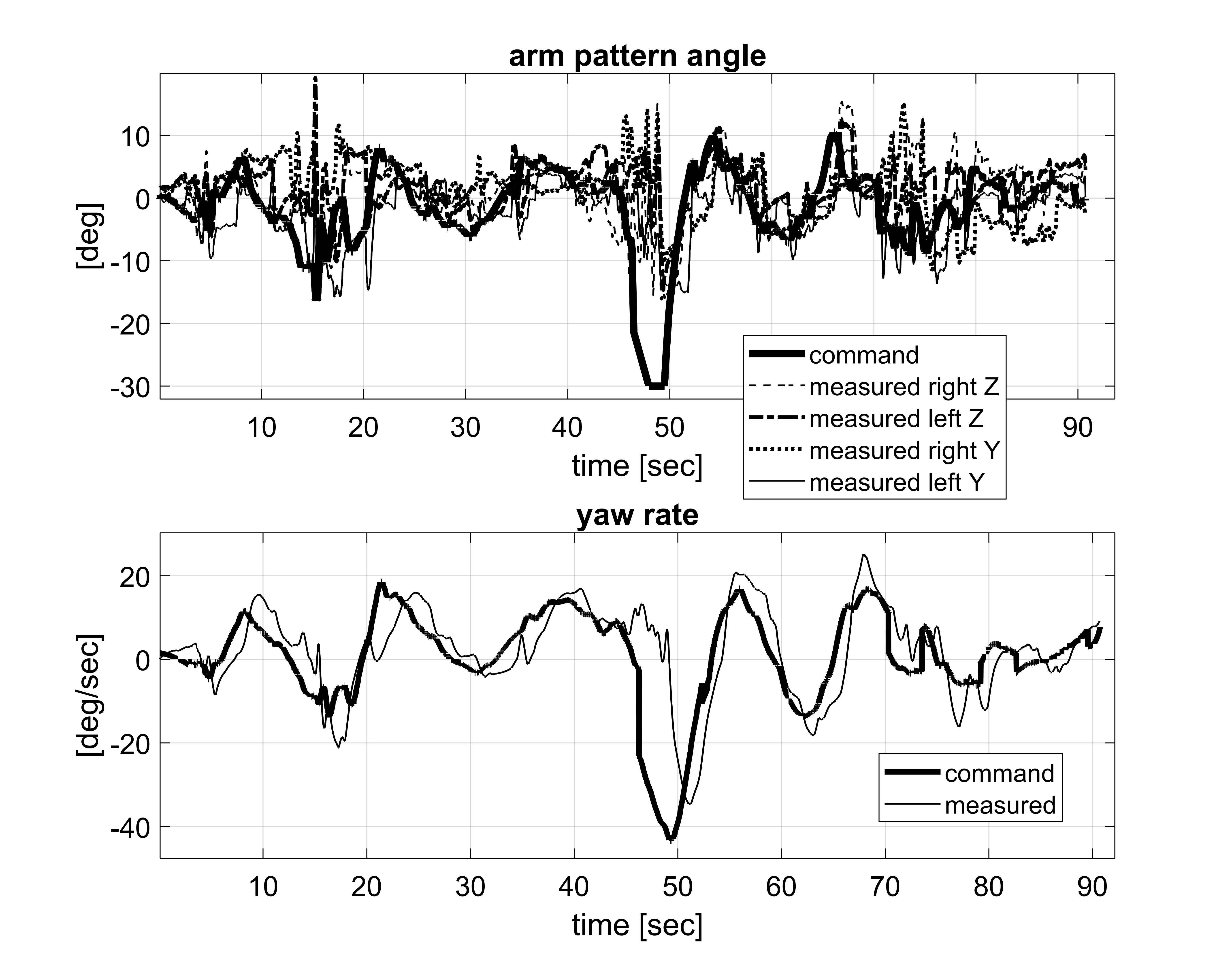}
    \caption{Posture and motion of the participant, who focused only on the Desired Posture cue, compared to the commands. The shoulder flexion and rotation degrees-of-freedom are denoted by Y and Z, respectively}
    \label{fig:person1}
\end{figure}

\begin{figure}[!t]
     \includegraphics[width=1\columnwidth]{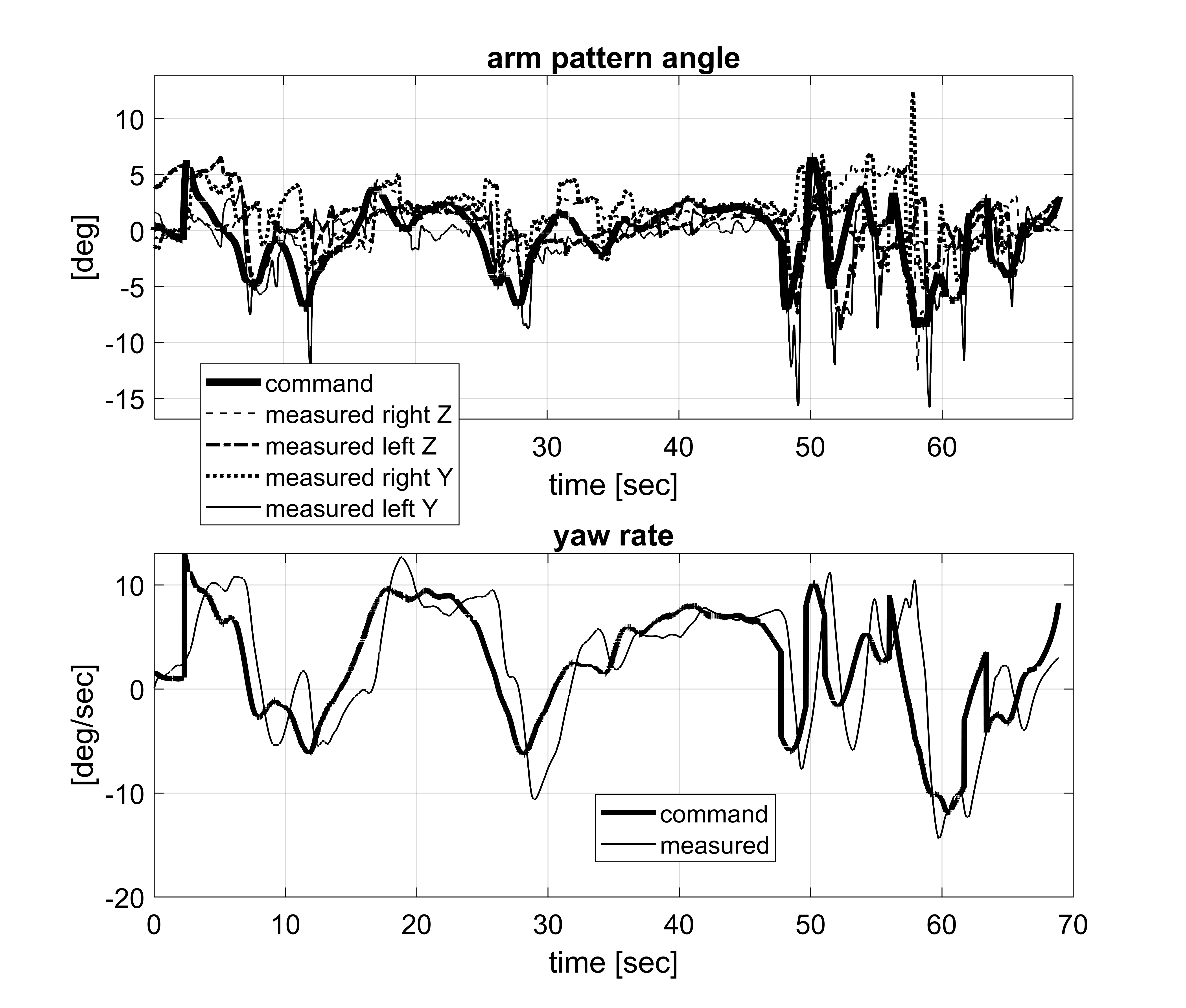}
    \caption{Posture and motion of the participant, who focused on the Desired Posture cue but also glanced at the Forward Model cue, compared to the commands. The shoulder flexion and rotation degrees-of-freedom are denoted by Y and Z, respectively}
    \label{fig:person2}
\end{figure}

\begin{figure}[!t]
     \includegraphics[width=1\columnwidth]{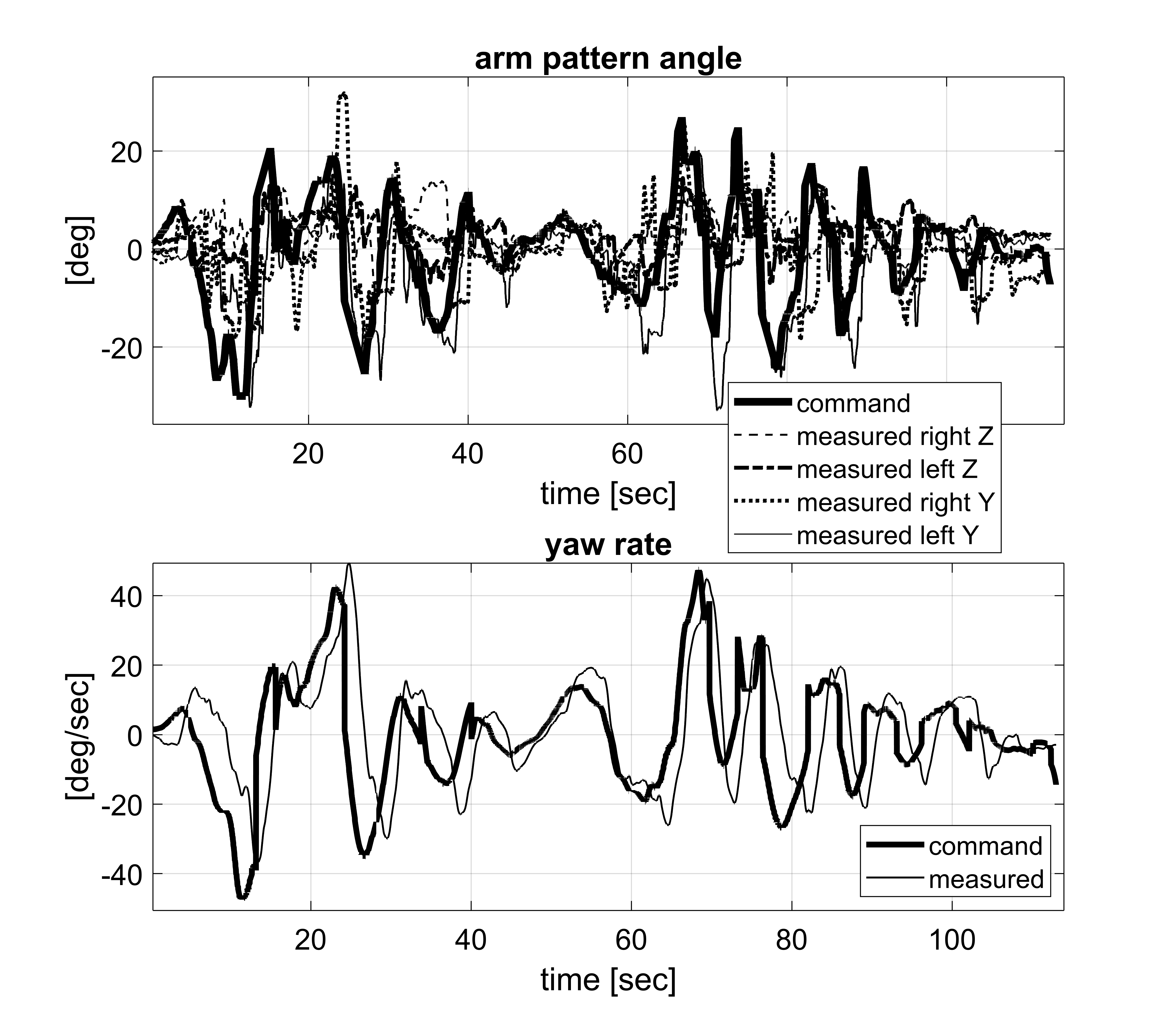}
    \caption{Posture and motion of the participant, who felt a delay while focusing on the Desired Posture cue, compared to the commands. The shoulder flexion and rotation degrees-of-freedom are denoted by Y and Z, respectively}
    \label{fig:person3}
\end{figure}

\begin{figure}[!t]
     \includegraphics[width=1\columnwidth]{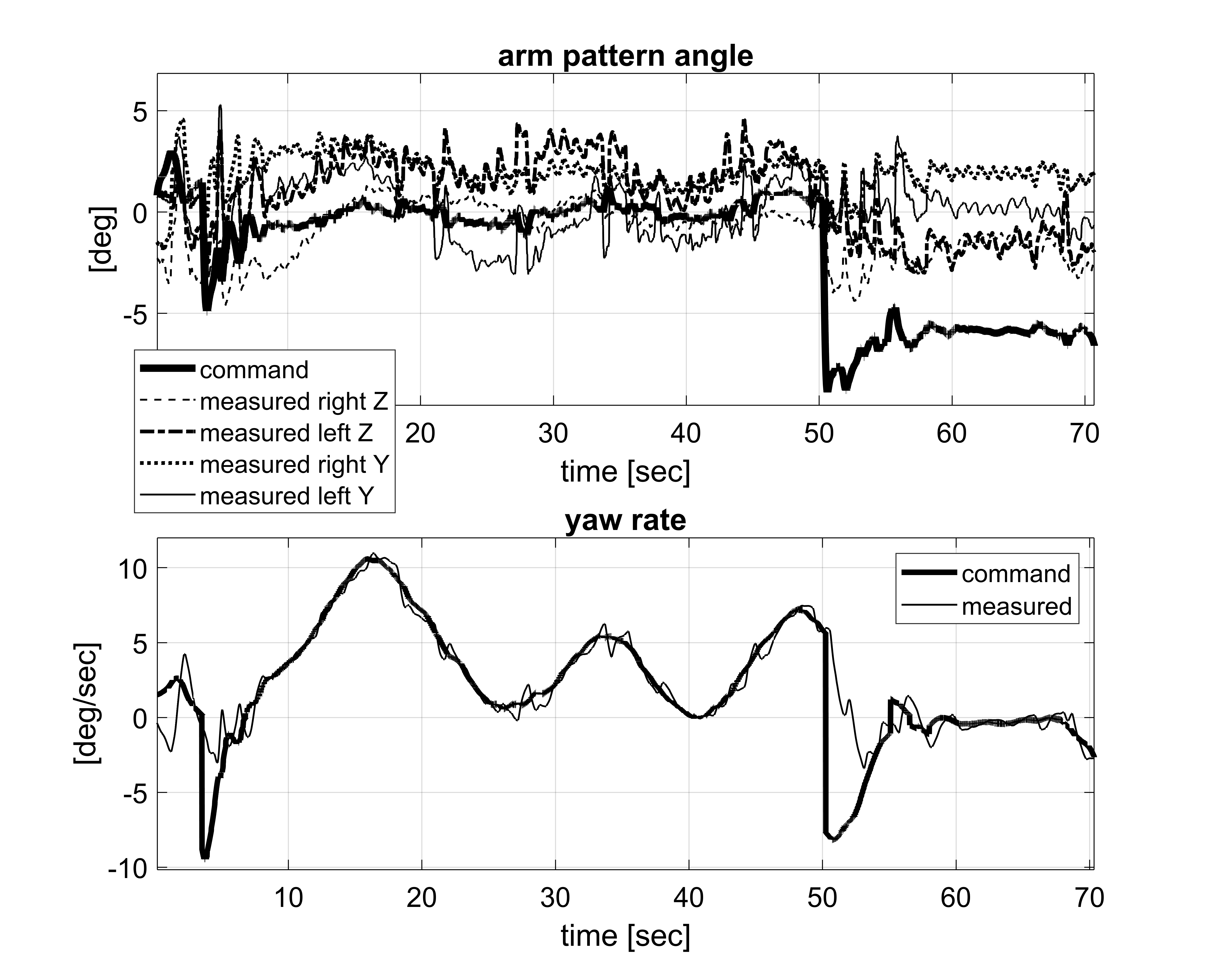}
    \caption{Posture and motion of the participant, who focused mostly on the Forward Model cue, compared to the commands. The shoulder flexion and rotation degrees-of-freedom are denoted by Y and Z, respectively}
    \label{fig:person4}
\end{figure}

Most important  is that the KTM cues greatly improved the participants' ability to fly the virtual skydiver. Figs. \ref{fig:group1_topview_free} and \ref{fig:group2_topview_free}  reveal that nobody succeeded to fly within the desired corridor  
without the KTM cues. After trying for a few minutes, all participants reported the task to be extremely hard and would probably take them hours or days of practice. However, after displaying the KTM cues, all the participants completed the task from the first attempt, within 1.5-2 minutes, see Figs. \ref{fig:group1_topview}, \ref{fig:group2_topview} and  \ref{fig:person1}-\ref{fig:person4}!  

As expected, see Section \ref{concept}, most  participants reported that they focused on the Desired Posture cue. Two participants succeeded in switching between the mainly-in-focus Desired Posture and the briefly-glanced Forward Model arrow cues, allowing them to continuously keep in mind the deviation from the desired orientation, while concentrating on decreasing it by the posture adjustment. These participants completed the task faster, followed the corridor more accurately, and had less gap between the actual and desired body posture and yaw rate,  for comparison see Figs. \ref{fig:person1}, \ref{fig:person2}.  Notice, that in Fig. \ref{fig:person2} the maximal required arm pattern angle command is 8~[deg], and the maximal yaw rate command is 12~[deg/sec], while the corresponding values from Fig.~\ref{fig:person1} are 30~[deg] and 42~[deg/sec], respectively. This means that the participant who didn't look at the Forward Model cue had to apply more control effort, as he didn't look at the preview of how much effort is required. Instead, he learnt this from the posture adjustments he had to make, fixing the disparity between desired and measured yaw rate. 

The participant with the most difficulty during the first attempt, 
reported that it took time to notice changes in the desired  posture. Predictably, delay in following the desired posture caused the control loop to exhibit oscillations, see Fig. \ref{fig:person3}. Nevertheless, he completed the task within 110~[sec], compared with 70-90~[sec] for the others. 

One of the participants focused mostly on the Forward Model cue, since she succeeded to acquire  muscle memory for the 'turning' pattern from the Imitation stage. She was more focused and spent more time imitating the pattern  than other participants. 
However, if one feels what the correct body movement is, 
it only remains to decide the movement amplitude, which is resolved by the Forward Model cue. The 'turning' pattern angle calculated by the KTM for this participant had the smallest variation over time among all participants, and the yaw rate command was followed most accurately, see Fig. \ref{fig:person4}. The maximal amplitude of the actually performed arm movement was  5~[deg], while it was in the range of 15-30~[deg] for the others.  

\section{Discussion}
There were two additional volunteers who could not participate for the following reasons:

\paragraph{Restricted rotation of the shoulder joint} One person had much less freedom of movement in the shoulders and could not repeat the movement required by the 'turning' pattern. This problem is known among older people and those not engaging the shoulder joints in every day activities. The solution is to offer multiple movement patterns that involve different DOFs but produce the same maneuver, as was discussed in Sec. \ref{concept}.  
\paragraph{Extremely unusual body dimensions} One volunteer was extremely tall (over 195 cm), and the neutral body pose offered by the simulator was highly unstable for him (caused backsliding). Also, the suggested 'turning' pattern caused more backsliding than the 'forward' pattern could compensate for. The compensation term is $G21$ in Eq. \ref{eq:loop1}. Skydiving instructors report that teaching tall people is challenging as they need to adopt completely different body postures than others. The KTM is the tool to resolve this problem: It can determine the neutral posture and design efficient movement patterns for any type/size of body. This should be done offline, by the means of Skydiving Simulator updated with the individual body parameters. For trainees with an intermediate level of skill, who already have an accustomed neutral posture, it is possible to measure this posture and use it in the KTM as $P_{neutral}$. The training system will be more effective if a neutral posture and movement patterns are designed individually and the controllers are tuned accordingly.

Based on the experiment results additional improvements for the KTM controllers can be suggested. Firstly, computation of the tracking errors ($\Omega_{com}(t)-\Omega_{meas}(t)$ and $V_{com}(t)-V_{meas}(t)$) can include prediction. Suppose, the trainee implements the desired posture with a certain delay, as in  Fig. \ref{fig:person3}. It is possible to estimate this delay time $t_{delay}$, solve the equations of motion to predict his angular and linear velocities in $t_{delay}$, and use them for calculating the tracking errors, which are the controllers' inputs: $\Omega_{com}(t)-\Omega_{meas}(t+t_{delay})$ and $V_{com}(t)-V_{meas}(t+t_{delay})$  

Secondly, in case of large gap between desired and executed postures, as in Fig. \ref{fig:person1} around time=50~[sec], adaptive control can be introduced. The simulation computes at each step what would be the angular and linear velocities if same velocity profiles are tracked by an ideal performer. The disparity between these velocities and the 'measured' ones is fed into classic PI controllers. Their outputs are added to the angles commands describing the desired movement patterns.

Adjusting the controllers to match the trainee's learning ability will help the initial stage of practice, when the Desired Posture cue is the most dominant. 

\section{Conclusions}
The Proof-of-Concept experiment, which required human subjects to control a virtual skydiver by the means of their body, has been conducted. The task was to fly along a pre-defined path.
This path was calculated by the navigation module, which plans the desired path and speed profile off-line, and then calculates on-line the desired yaw rate. The yaw rate and speed profiles were tracked by the two controllers that output the angle commands for the two corresponding movement patterns. The movement patterns define the desired body posture. Every level of the control hierarchy was represented in the cues displayed to the trainees: the desired path, velocities, and posture. The cues proved to be very efficient, since all the participants completed the task from the first attempt.

This means that the trainees acted as a part of the closed control loop tracking the desired path, where they fulfilled the role of an actuator. Their body movements were continuously guided by the visual cues without sensing the actual motion and in the absence of knowledge about the free-fall dynamics. This system with human-in-the-loop was stable and enabled trainees to control the virtual skydiver, thus indicating the success of the Proof-of-Concept experiment. The hypotheses of the interface between the human and automatic control parts of the system  were also verified:    
At the beginning of the practice the Desired Posture cue was the most dominant. The Forward Model cue became dominant when trainees were able to perform the pattern from muscle memory, i.e. without looking at the posture cue. At this stage, trainees learn to adjust the amplitude of the pattern to the task, using the Forward Model cue as a feedback. Next, a variety of tasks should be introduced to the trainees, so that they get acquainted with the maneuver range, that can be achieved by the trained pattern. Next step after that can be performing the same tasks by the means of different movement patterns. Designing an efficient training program and building a KTM prototype will be the future work.
\ifCLASSOPTIONcaptionsoff
  \newpage
\fi



\bibliographystyle{IEEEtran}
\end{document}